\documentclass{article}

    \PassOptionsToPackage{numbers, compress}{natbib}

\usepackage[preprint]{neurips_2026}

\makeatletter

\@ifpackageloaded{inputenc}{}{\usepackage[utf8]{inputenc}}
\@ifpackageloaded{fontenc}{}{\usepackage[T1]{fontenc}}
\@ifpackageloaded{natbib}{}{\usepackage[numbers]{natbib}}
\@ifpackageloaded{url}{}{\usepackage[hyphens]{url}}
\@ifpackageloaded{hyperref}{}{\usepackage{hyperref}}

\@ifpackageloaded{booktabs}{}{\usepackage{booktabs}}
\@ifpackageloaded{threeparttable}{}{\usepackage{threeparttable}}
\@ifpackageloaded{makecell}{}{\usepackage{makecell}}
\@ifpackageloaded{array}{}{\usepackage{array}}
\@ifpackageloaded{rotating}{}{\usepackage{rotating}}
\@ifpackageloaded{multirow}{}{\usepackage{multirow}}
\@ifpackageloaded{multicol}{}{\usepackage{multicol}}

\@ifpackageloaded{amsfonts}{}{\usepackage{amsfonts}}
\@ifpackageloaded{amsmath}{}{\usepackage{amsmath, amsthm, amssymb, mathtools}}
\@ifpackageloaded{nicefrac}{}{\usepackage{nicefrac}}

\@ifpackageloaded{graphicx}{}{\usepackage{graphicx}}
\@ifpackageloaded{subfigure}{}{\usepackage{subfigure}}
\@ifpackageloaded{wrapfig}{}{\usepackage{wrapfig}}

\@ifpackageloaded{algorithm}{}{\usepackage{algorithm}}
\@ifpackageloaded{algorithmicx}{}{\usepackage{algorithmicx}}
\@ifpackageloaded{algpseudocode}{}{\usepackage{algpseudocode}}

\@ifpackageloaded{microtype}{}{\usepackage{microtype}}
\@ifpackageloaded{xcolor}{}{\usepackage[table]{xcolor}}
\@ifpackageloaded{tcolorbox}{}{\usepackage[most]{tcolorbox}}
\@ifpackageloaded{adjustbox}{}{\usepackage{adjustbox}}
\@ifpackageloaded{enumitem}{}{\usepackage[inline]{enumitem}}
\@ifpackageloaded{placeins}{}{\usepackage{placeins}}
\@ifpackageloaded{inconsolata}{}{\usepackage{inconsolata}}
\@ifpackageloaded{cleveref}{}{\usepackage[capitalize,noabbrev]{cleveref}}

\@ifpackageloaded{pgfplots}{}{\usepackage{pgfplots}\pgfplotsset{compat=1.17}}

\makeatother

\hypersetup{
  hidelinks,
}

\newcommand{\email}[2][@yale.edu]{\href{mailto:#2#1}{\texttt{#2}}}

\definecolor{gemmacolor}{HTML}{1a8c1a}
\definecolor{llamacolor}{HTML}{cc6600}

\newtcbox{\highlight}[1][customorange]{
    on line,
    arc=0pt,
    colback=#1!35!white,
    colframe=#1!35!white,
    boxsep=0pt,
    left=1pt,
    right=1pt,
    top=2pt,
    bottom=2pt,
    boxrule=0pt,
    nobeforeafter
}

\newcommand{\best}[1]{{\color{black!80}\textbf{#1}}}

\usepackage{pifont}

\newcommand{\topcap}[1]{\begin{tabular}[t]{@{}c@{}}#1\end{tabular}}



\title{Test-Time Safety Alignment}

\author{%
  Baturay Saglam \quad Dionysis Kalogerias \\ \\
  Department of Electrical and Computer Engineering\\
  Yale University\\ \\
  \{\small \email{baturay.saglam}, \email{dionysis.kalogerias}\}\texttt{@yale.edu}
}

\begin{document}

\maketitle

\begin{abstract}
    Recent work has shown that a model's input word embeddings can serve as effective control variables for steering its behavior toward outputs that satisfy desired properties. However, this has only been demonstrated for pretrained text-completion models on the relatively simple objective of reducing surface-level profanity in short continuations. A natural and practically important question is how well input embeddings can control aligned models, which produce an imbalanced bimodal refuse-or-comply output distribution rather than the smooth distribution characteristic of open-ended generation. We explore this in the context of safety, showing that input word embeddings can be optimized in a sub-lexical manner to minimize the semantic harmfulness of aligned model responses. Our approach uses zeroth-order gradient estimation of a black-box text-moderation API with respect to the input embeddings, and then applies gradient descent on these embeddings to minimize the harmfulness of the generated text. Experiments show that the proposed method can neutralize every safety-flagged response on standard safety benchmarks.
\end{abstract}

\section{Introduction}
\label{sec:introduction}

Large language models deployed for user-facing tasks undergo safety training---typically reinforcement learning from human feedback (RLHF)~\citep{rlhf, harmless-assistant} and direct preference optimization~\citep{dpo}---to handle requests that violate their developers' safety policies. Despite this training, adversarial prompts can reliably bypass safety mechanisms and elicit harmful content from otherwise well-aligned models~\citep{gcg, pair}. These prompts wrap harmful requests in structural framing---roleplay scenarios, hypothetical narratives, obfuscating instructions---that exploits the gap between the model's safety training distribution and the space of possible inputs. The vulnerability is fundamental: alignment is learned from a finite set of examples, while the input space is effectively unbounded.

A growing body of test-time defenses addresses this gap without modifying the target model's weights. Prompt-level strategies~\citep{self-reminder, icd, rpo}, randomized smoothing~\citep{smoothllm, semantic-smooth}, guard models~\citep{llama-guard, wildguard}, and activation-steering methods~\citep{adasteer, abd} have all demonstrated meaningful reductions in attack success rates. Yet these approaches share a structural limitation: they either apply undirected perturbations to discrete tokens, require separately offline-trained auxiliary modules, or impose fixed interventions derived from a static calibration set. None directly optimizes a continuous safety objective at the model's input level.

A qualitatively different approach is to operate on the continuous representation the model actually conditions on: the prompt embedding matrix. Because the model reads this matrix rather than the tokens themselves, perturbations in embedding space can steer the output distribution without altering the human-readable input. \citet{tide} demonstrated this principle for detoxification, showing that iterative optimization over prompt embeddings can reduce the toxicity of base pretrained models as scored by a surface-level profanity classifier. Whether embedding-level control extends to the safety of aligned models is, however, an open question---and there is reason to expect that it may not transfer straightforwardly. Aligned models produce a bimodal output distribution---refuse or comply---rather than the smooth continuation distribution over which toxicity reduction was demonstrated; the relevant notion of harm is semantic (violence, self-harm, illicit activity) rather than lexical (profanity, slurs); and the prompts of interest are adversarially constructed to circumvent safety training, a qualitatively different regime from the naturally occurring toxic prompts studied previously.

We show that the transfer does hold. Given any prompt, we apply zeroth-order gradient estimation to approximate the gradient of a black-box text-moderation oracle with respect to the prompt embeddings, and descend along the estimated gradient to minimize the harmfulness of the generated text. The oracle—the OpenAI Moderation API~\citep{openai_moderation, openai_moderation_url}—scores completions across 13 harm categories; we take the maximum over these categories as our objective, so that a model response is considered safe only if it scores low along every axis of harm. The entire procedure operates at test time: it requires no retraining and no auxiliary models, and treats the language model as a black box—only its input word embeddings need to be accessible. The optimization is applied unconditionally, without knowledge of whether a prompt is harmful or benign, and does not force refusals; the model determines the form of its own response.

Experiments on five instruction-tuned models spanning 1B to 20B parameters, evaluated on two standard red-teaming benchmarks comprising adversarial jailbreak prompts and direct harmful queries, show that the method nearly eliminates every safety-flagged completion, typically converging within two gradient steps. On an adversarial benign control split, already-safe responses are left substantively intact---moderation scores do not increase for any model---confirming that the optimization does not impose a blanket refusal bias. A finding of broader interest is that the optimized embeddings decode back to the original prompt tokens in every case: the perturbation is sub-lexical, adjusting the continuous representation within the nearest-neighbor cell of each token, yet it reliably redirects the model toward safe outputs. This observation suggests that the conventional view of embeddings as inert token lookups understates the \emph{control surface} available at the model's input layer. Our codebase includes a demo notebook illustrating the proposed method in action.\footnote{\url{https://github.com/baturaysaglam/instant-alignment}}

\section{Related Work}
\label{sec:related_work}

We review inference-time methods for defending aligned language models without modifying the target model's weights. Approaches that improve safety through training~\citep{constitutional-ai} or fine-tuning~\citep{rlhf,harmless-assistant,dpo} are complementary but outside our scope.

\paragraph{Prompt-level and input-transformation defenses.}
The earliest test-time defenses operate on the discrete text the model receives. System-prompt strategies wrap user queries with safety reminders~\citep{self-reminder} or prepend in-context refusal demonstrations~\citep{icd}, while more principled variants optimize defensive suffixes offline through minimax adversarial objectives~\citep{rpo} or adversarial prompt training~\citep{pat} and deploy the result as a fixed textual guard.
%
%
A complementary direction randomizes the input to disrupt adversarial structure~\citep{smoothllm,semantic-smooth} or targets systematic token erasure~\citep{erase-and-check}. All of these methods manipulate discrete tokens or strings. Our method instead operates in the continuous embedding space and employs directed gradient-based optimization rather than undirected perturbation or fixed textual controls.

\paragraph{Safety classifiers and detection.}
A second line of work interposes auxiliary models or analyses between the user and the LLM. Llama Guard~\citep{llama-guard} and WildGuard~\citep{wildguard} finetune language models into safety classifiers that filter harmful inputs or outputs at deployment.
%
%
Other methods extract detection signals directly from the target model, such as perplexity-based filtering~\citep{perplexity-detection}, gradient-based methods for pattern identification~\citep{gradsafe} and refusal-loss thresholding~\citep{gradient-cuff}, or linear probing for jailbreak detection~\citep{jbshield}. These approaches share a detect-then-filter paradigm: they classify the input and block or reclassify it. Our method instead steers generation fully autonomously in an input-driven manner, without having to predict intent and therefore without requiring a separate classification model or detection step.

\paragraph{Activation steering and decoding-time interventions.}
A third family intervenes on the model's internal computation during inference. Activation-steering methods add safety-relevant directions to hidden states---either as fixed vectors transferred from an aligned donor model~\citep{inferaligner}, with per-input adaptive coefficients derived from a calibration set~\citep{adasteer}, or by constraining activations to lie within empirically estimated safety boundaries~\citep{abd}.
%
%
At the decoding level, SafeDecoding~\citep{safe_decoding} contrasts the output distributions of the frozen model and a lightweight safety expert to amplify safe tokens, and RAIN~\citep{rain} integrates self-evaluation and rewind mechanisms during generation. These approaches require access to hidden states or output logits and, in the case of activation steering, derive directions from a fixed calibration set that must transfer across prompts. Our method instead treats the model as a black-box system—assuming only that the input word embeddings are available—and, thanks to its black-box optimization, adapts its intervention at the instance level, targeting only the prompt at hand.

\paragraph{Embedding-space methods.}
The work most closely related to ours operates at the embedding level. RESTA~\citep{resta} injects isotropic Gaussian noise into token embeddings and aggregates multiple perturbed copies via majority vote—randomized smoothing in embedding space, with no directed optimization and therefore no ability to target a specific safety objective as we do. DRO~\citep{dro} optimizes continuous soft-prompt embeddings to shift representations toward higher-refusal regions, achieving directed control but requiring an offline training phase and producing a fixed prompt that is applied uniformly at deployment. In contrast, we optimize the embeddings against a semantic moderation objective rather than perturbing them at random, and we apply this optimization directly to the word embeddings of the prompt at hand, fully at test time, without any offline training or additional data. The algorithmic foundation of our approach is based on \citep{tide}, which showed that iterative optimization over prompt embeddings can reduce the toxicity of base pretrained (non–instruction-tuned) models in open-ended generation, using a surface-level profanity scorer. We apply this framework to a fundamentally different regime: safety-aligned chat models, a multi-category semantic moderation oracle, and adversarial or malicious prompts designed to circumvent safety training through semantic manipulation, rather than naturally occurring short, toxic continuations.

\section{Problem Formulation}
\label{sec:problem-formulation}

Modern large language models intended for user-facing deployment are \emph{aligned}: in addition to next-token pretraining on web-scale corpora, they undergo a post-training stage designed to make their outputs helpful, honest, and harmless. Alignment is typically achieved through some combination of supervised instruction fine-tuning~\citep{ouyang_ift}, RLHF \citep{rlhf, harmless-assistant}, and direct preference optimization \citep{dpo}. The resulting models are capable of following natural-language instructions across a wide range of tasks—answering questions, writing code—while refusing requests that violate the safety policies of their developers, such as instructions to facilitate violence or assist in illicit activity.

Let $f$ denote an aligned autoregressive language model based on the transformer architecture \citep{transformer}. A prompt is a string $\mathcal{P}$ that a tokenizer maps to a sequence of token indices
\[
t_{1:T} = (t_1, \ldots, t_T), \qquad t_i \in \mathcal{V},
\]
drawn from a finite vocabulary $\mathcal{V}$. Each token $t_i$ is associated with a fixed $d$-dimensional embedding vector $x_i \in \mathbb{R}^d$, so the full prompt is represented by the embedding matrix
\[
X = \begin{bmatrix} x_1^\top & x_2^\top & \cdots & x_T^\top \end{bmatrix}^\top \in \mathbb{R}^{T \times d}.
\]
Given a prefix $(t_1, \ldots, t_{i-1})$, the model defines a distribution over the next token,
\[
f(t_i \mid t_1, \ldots, t_{i-1}) \;\equiv\; p_f(t_i \mid t_{<i}).
\]
We denote by $y = f(X)$ the response obtained by sequentially sampling from this distribution when $f$ is conditioned on the prompt $\mathcal{P}$---equivalently, on its token sequence $t_{1:T}$ and embedding matrix $X$.



\subsection{Measuring Harmful Generations}

Despite alignment, models can still produce harmful content—whether through residual training gaps, distribution shift, or adversarial prompts specifically designed to elicit such responses. This content may carry no surface-level markers: a response can be polite and profanity-free yet provide detailed instructions for illicit activity. Detecting such failures requires evaluating the \emph{generated response}, not the prompt.

We measure the harmfulness of $y = f(X)$ using the OpenAI Moderation API~\citep{openai_moderation, openai_moderation_url}, a black-box content classifier deployed so that hosts and users can take corrective action—filtering outputs, flagging accounts, or blocking responses—when potentially harmful content is detected. The API returns a continuous score in $[0, 1]$ for each of 13 categories: harassment, harassment/threatening, hate, hate/threatening, illicit, illicit/violent, self-harm, self-harm/intent, self-harm/instructions, sexual, sexual/minors, violence, and violence/graphic. Each category additionally carries a binary ``flagged'' indicator; the threshold mapping scores to flags is internal to the API and not exposed. We abstract this oracle as
\begin{equation}
    \label{eq:moderation-max}
    h: \mathcal{Y} \to [0, 1], \qquad h(y) = \max_{c}\, s_c(y),
\end{equation}
where $\mathcal{Y}$ is the space of output strings and $s_c(y)$ is the score in category $c$. Taking the maximum ensures that a response is considered safe only if it scores low across \emph{every} harm category---minimizing $h$ therefore always targets the worst-offending category.

\subsection{Objective}

Our goal is to intervene at test time—without modifying the target model or learning any additional parameters—so that the harmfulness of the model's output is minimized for any prompt, regardless of its intent or construction. The method is applied unconditionally: it has no access to whether a prompt is benign or harmful, and it does not force refusals—the model determines the form of its own response. Concretely, we optimize the prompt embedding matrix $X$ to reduce the harmfulness of the resulting response $\Phi(X) \coloneqq h\left(f(X)\right)$, treating both the language model $f$ and the moderation oracle $h$ as black boxes. Because backpropagating through $f$ is prohibitively expensive at the scale of modern LLMs, we restrict ourselves to forward passes alone.

Since minimizing $h$ already targets the worst-offending category at every step, the result is a fully automated mechanism that steers the model toward safe responses precisely when caution is warranted, while leaving already-safe responses substantively intact.

\section{Methodology}
\label{sec:methodology}

\paragraph{Overview.}
The composite objective $\Phi(X) = h\left(f(X)\right)$ exposes no usable gradient; $f$ is an autoregressive generator and $h$ is a black-box cloud API. We therefore substitute a stochastic surrogate: at iteration $k$ we form an estimator $g_k$—derived in the next subsection from a Gaussian-smoothed reformulation of $\Phi$—and apply the update
\[
X_{k+1} \leftarrow X_k - \eta\, g_k,
\]
which approximately decreases $\Phi$ and thereby drives the resulting response toward lower-harmfulness regions of embedding space, typically converging within a handful of iterations (in practice, $K < 4$).

\subsection{Zeroth-Order Gradient Estimator}

Unless noted otherwise, the development in this subsection follows the foundation in \citep{nesterov_zoo}. Our objective is to estimate $\nabla_X \Phi(X)$ given only forward (zeroth-order) access to $\Phi$.

We begin by perturbing $X$ with i.i.d.\ Gaussian noise applied independently to each token embedding:
\[
X + \mu U, \qquad
U =
\begin{bmatrix}
u_1^\top & u_2^\top & \cdots & u_T^\top
\end{bmatrix}^\top \in \mathbb{R}^{T \times d},
\qquad
u_1, \ldots, u_T \overset{\text{i.i.d.}}{\sim} \mathcal{N}(0, I_d).
\]
Since $U$ has the same shape as $X$, every token is perturbed in its own direction, allowing the resulting estimator to capture token-specific sensitivities of $\Phi$ and to adjust the prompt at a fine-grained level. Averaging $\Phi$ over these perturbations defines the \emph{Gaussian-smoothed} objective
\[
\Phi_\mu(X) = \mathbb{E}_{U}\!\left[\Phi(X + \mu U)\right].
\]
Smoothing has two convenient consequences. First, whenever $\Phi$ is Lipschitz-continuous—i.e., $|\Phi(X) - \Phi(Y)| \le L\|X - Y\|$—the smoothed surrogate $\Phi_\mu$ is differentiable for every $\mu > 0$ while remaining Lipschitz itself. Second, replacing $\Phi$ with $\Phi_\mu$ incurs only a controlled approximation error of order $\mathcal{O}(\mu\sqrt{Td})$.

A key result of \citet{nesterov_zoo} expresses the gradient of the smoothed objective in closed form as a directional finite difference:
\begin{equation}
\label{eq:grad_smoothed_obj}
\nabla\Phi_\mu(X) = \mathbb{E}_{U}\!\left[
\frac{\Phi(X + \mu U) - \Phi(X)}{\mu}\, U
\right].
\end{equation}
This identity requires only Lipschitz continuity of $\Phi$; differentiability of $\Phi$ itself is not assumed. Intuitively, $\nabla\Phi_\mu(X)$ is a weighted average of perturbation directions, each scaled by the relative change $\Phi$ exhibits along it. The subtracted baseline $\mathbb{E}_U[\Phi(X)\,U/\mu]$ leaves the expectation unchanged, since $\mathbb{E}_U[U] = 0$ and $\Phi(X)$ is independent of $U$, yet it reduces the variance of any Monte Carlo approximation.

To relate $\nabla\Phi_\mu$ back to $\nabla\Phi$, one additional regularity assumption is useful. If $\Phi$ is also Lipschitz-smooth—that is, if its gradient is itself $L$-Lipschitz—then $\Phi_\mu$ inherits the same smoothness, and the gap between the two gradients is bounded:
\[
\|\nabla\Phi_\mu(X) - \nabla\Phi(X)\| \le C\,\mu L\,(Td)^{3/2},
\]
for a constant $C$. Lipschitz continuity stabilizes the finite difference, while smoothness controls how rapidly $\nabla\Phi_\mu$ itself can vary. Together they make $\nabla\Phi_\mu(X_k)$ a controlled approximation of $\nabla\Phi(X_k)$, with $\mu$ governing the residual bias. We stress that these conditions are invoked only to motivate the bound; in a fully black-box setting they cannot be verified from query access to $\Phi$. The method itself requires nothing beyond forward evaluations and access to input embeddings, and can therefore be applied empirically regardless of whether the underlying constants are known.

We should note that the maximum in~\eqref{eq:moderation-max} is not differentiable, so $\Phi$ may be Lipschitz-continuous but not Lipschitz-smooth, and therefore the bound above does not apply as stated. A smooth surrogate---for instance, a softmax-weighted average $\sum_c w_c\, s_c$ with weights $w_c \propto \exp(\beta\, s_c)$---would restore the required regularity. In practice, however, we omit this substitution; across our experiments the highest-scoring harm category remains stable throughout optimization for each prompt, so the maximum effectively reduces to a single smooth category score along the optimization path.

Finally, we replace the expectation in \eqref{eq:grad_smoothed_obj} with a Monte Carlo average over $N$ independent perturbations, yielding the practical estimator used throughout this work:
\begin{equation*}
\boxed{
g_k = \frac{1}{N} \sum_{i=1}^{N}
\frac{\Phi(X_k + \mu U_i) - \Phi(X_k)}{\mu}\, U_i
\;\approx\; \nabla\Phi_\mu(X_k).
}
\end{equation*}
The estimator sits on top of ordinary model inference and requires no changes to the model internals. Its variance grows with the ambient dimension $Td$ and shrinks with $N$, while $\mu$ controls a bias--variance tradeoff. Larger $\mu$ smooths $\Phi$ more aggressively and stabilizes the estimate, whereas smaller $\mu$ sharpens the approximation at the cost of noisier evaluations. In the joint limit $\mu \to 0$ and $N \to \infty$, and under the regularity conditions discussed above, $g_k$ recovers the true gradient direction of $\Phi$; in practice, moderate values of both suffice for stable and effective test-time updates.

\subsection{Regularization}

The zeroth-order descent procedure described so far must be regularized to keep the updated embeddings from drifting into regions of the latent space that the model has not encountered during training. We use a few deliberately simple mechanisms and introduce no auxiliary modules—no momentum, no learned projection, no external reward shaping. This isolates the effect of the zeroth-order updates themselves, so that any improvement we observe is attributable to the test-time intervention alone.

\paragraph{Gradient normalization.}
%
A single fixed learning rate applied to the raw estimate can produce erratic updates—some too timid to make progress, others large enough to push embeddings off-manifold. We sidestep this by normalizing the estimate before each step, replacing $g_k$ with $g_k / \|g_k\|_2$ in the update rule. The effective step magnitude is then governed entirely by $\eta$, yielding a single learning rate that behaves consistently across prompts.

\paragraph{Cosine similarity.}
To preserve fidelity to the original prompt, we additionally constrain the updated embedding to remain close to $X_0$ in direction. Cosine similarity is well suited here because it measures alignment independently of magnitude. At the end of each iteration, if the current embedding falls below a cosine similarity threshold $\kappa$ relative to $X_0$, we project it back onto the boundary of the corresponding cosine ball before proceeding.

We refer to the resulting end-to-end test-time defense as \textbf{T}est-Time \textbf{S}afety \textbf{A}lignment (TSA); pseudocode is provided in Algorithm~\ref{alg:tsa} (Appendix~\ref{app:pseudocode}).
\section{Experiments}
\label{sec:experiments}

\paragraph{Models.}
Experiments are conducted on several aligned, instruction-tuned models from a range of developers and across several scales: Gemma 3 (1B)~\citep{gemma3}, Phi-3.5-mini (4B)~\citep{phi3.5}, Llama 3.1 (8B)~\citep{llama3}, Qwen3 (14B)~\citep{qwen3}, and GPT-OSS (20B)~\citep{gpt-oss}.

\paragraph{Benchmarks.}
To elicit responses that contain harmful information, we select red-teaming benchmarks where test prompts vary along two independent axes: \emph{construction} (direct natural-language phrasing vs.\ adversarial framing such as roleplay scenarios or obfuscating instructions) and \emph{intent} (benign vs.\ harmful). These axes are independent: a direct prompt may be harmful (an explicit request for dangerous information), and an adversarial prompt may be benign (a creative writing exercise with unusual framing). Adversarially framed harmful prompts constitute the most challenging condition, since their structural framing (i.e., a prompt injection) is specifically designed to elicit harmful content from aligned models.

For adversarial prompts, we use the evaluation set of the WildJailbreak benchmark~\citep{wildjb}, which contains 210 adversarial benign prompts (as a sanity check) and 2{,}000 adversarial harmful prompts. We also include HarmBench~\citep{harmbench}, which provides 400 direct harmful queries. Details about the benchmark datasets are provided in Appendix~\ref{app:datasets}.

\paragraph{Generation.}
Models are queried over three trials at temperature 0.1, generating 128 tokens per response. A budget of 128 tokens is sufficient because the moderation oracle is saturated by any harmful span within the window. Given the safety-sensitive nature of the prompts, a low temperature concentrates sampling on the modal completion that a typical user would receive in deployment, making it a realistic regime in which to measure output harmfulness.

\paragraph{Metrics.}
We report the mean, median, and maximum of the highest-category score, averaged over the three completions. These continuous scores are accompanied by the number of \emph{flagged} (i.e., classified as dangerous) scenarios. The classifier remains a black box but is believed to balance precision and recall~\citep{openai_moderation}. The optimization runs for 10 iterations with early stopping triggered when the top-scoring category falls below 0.1. This threshold is chosen because the developers have found moderation scores above 0.1 to be actionable~\citep{moderation-th-devs}. Error bars are omitted since the magnitude of the proposed method's effects far exceeds any plausible variance from stochastic generation, as our experiments will show.

\paragraph{Baselines.}
We compare against three inference-time defenses: SmoothLLM~\citep{smoothllm}, AdaSteer~\citep{adasteer}, and RESTA~\citep{resta}. SmoothLLM randomly perturbs copies of the input at the character level and returns the majority-safe completion; AdaSteer adaptively steers hidden-state activations along safety-relevant directions derived from a small calibration set; and RESTA injects Gaussian noise into the prompt embeddings and aggregates over the perturbed completions. Baseline comparisons are conducted on Llama~3.1-8B-Instruct as it is the only model shared between our evaluations and the original papers. This choice allows the published configurations to be applied directly, avoiding the confounds of the independent per-model hyperparameter search that AdaSteer and RESTA would otherwise require due to their method-specific calibration parameters.

To our knowledge, no other test-time defense directly optimizes a continuous safety objective; these baselines represent the closest available inference-time alternatives and ensure an apples-to-apples comparison.

Further experimental details are provided in Appendix~\ref{app:experimental-details}.

\subsection{Main Results}

\begin{table*}[htp]
  \caption{Text moderation results for five instruction-tuned language models on two red-teaming benchmarks. Responses are generated at a sampling temperature of 0.1 with three independent trials per prompt. Scores report statistics of the maximum over the 13 OpenAI Moderation API harm categories per prompt, in $[0,1]$; lower is safer. Optimization runs for up to 10 iterations or until the maximum score across the 13 categories falls below 0.1. \emph{Flagged} counts the prompts whose completions the API classifies as harmful. \emph{Time} reports the average per-prompt optimization wall-clock duration, and \emph{Net} excludes moderation API latency. Within each (model, benchmark) pair, the best value for each metric is shown in \best{bold}.}
  \label{tab:main-results}
  
  \begin{center}
    \small
    \setlength{\tabcolsep}{6pt}
    \renewcommand{\arraystretch}{1.05}
    
    \begin{tabular}{%
      @{}
      c                             
      @{\hspace{16pt}}
      l                             
      r                             
      @{}p{20pt}@{}                  
      r                             
      r                             
      r                             
      @{}p{12pt}@{}                  
      r                             
      r                             
      r                             
      @{}
    }
    
    \toprule
      & & & & \multicolumn{3}{@{}c@{}}{Score\,$\downarrow$} & & & & \\
      \cmidrule(lr){5-7}
      & \textbf{Model} & Flagged\,$\downarrow$ & & Mean & Med. & Max & & \# Iterations & Time\,(s) & Net\,(s) \\
    \midrule
   
    \multirow{13}{*}[-10pt]{\rotatebox[origin=c]{90}{%
      \shortstack[c]{WildJailbreak\\[-0.25pt](Adv.\ Harmful)\\[-0.25pt]$n = 2{,}000$}}}
      & GPT-OSS-20B     &  17 && 0.023 & 0.004 & 0.849 && --  & --   & --   \\
      & \quad+\,TSA     &   \best{0} && \best{0.009} & \best{0.003} & \best{0.099} && 1.2 & 12.1 & 10.2 \\
      \addlinespace[5pt]
      & Qwen3-14B       & 165 && 0.097 & 0.028 & 0.944 && --  & --   & --   \\
      & \quad+\,TSA     &   \best{0} && \best{0.023} & \best{0.016} & \best{0.100} && 2.1 & 28.3 & 25.4 \\
      \addlinespace[5pt]
      & Llama 3.1-8B    & 218 && 0.106 & 0.020 & 0.951 && --  & --   & --   \\
      & \quad+\,SmoothLLM & 122 && 0.068 & 0.006 & 0.954 && --  & --   & --   \\
      & \quad+\,AdaSteer  &  85 && 0.057 & 0.016 & 0.947 && --  & --   & --   \\
      & \quad+\,RESTA     & 199 && 0.098 & 0.010 & 0.944 && --  & --   & --   \\
      & \quad+\,TSA     &   \best{0} && \best{0.012} & \best{0.002} & \best{0.098} && 1.0 & 12.2 & 10.5 \\
      \addlinespace[5pt]
      & Phi-3.5-4B      & 126 && 0.081 & 0.020 & 0.921 && --  & --   & --   \\
      & \quad+\,TSA     &   \best{0} && \best{0.019} & \best{0.008} & \best{0.099} && 1.8 & 12.4 &  9.8 \\
      \addlinespace[5pt]
      & Gemma 3-1B      & 278 && 0.137 & 0.063 & 0.946 && --  & --   & --   \\
      & \quad+\,TSA     &  \best{59} && \best{0.065} & \best{0.031} & \best{0.844} && 5.7 & 21.1 & 13.7 \\
    
    \midrule
    
    \multirow{13}{*}[-10pt]{\rotatebox[origin=c]{90}{%
      \shortstack[c]{WildJailbreak\\[-0.25pt](Adv.\ Benign)\\[-0.25pt]$n = 210$}}}
      & GPT-OSS-20B     &   \best{0} && 0.011 & \best{0.003} & 0.270 && --  & --   & --   \\
      & \quad+\,TSA     &   \best{0} && \best{0.009} & \best{0.003} & \best{0.097} && 1.0 &  8.0 &  6.3 \\
      \addlinespace[5pt]
      & Qwen3-14B       &   2 && 0.014 & \best{0.001} & 0.253 && --  & --   & --   \\
      & \quad+\,TSA     &   \best{0} && \best{0.010} & \best{0.001} & \best{0.098} && 2.1 & 29.2 & 26.3 \\
      \addlinespace[5pt]
      & Llama 3.1-8B    &   1 && 0.017 & 0.003 & 0.356 && --  & --   & --   \\
      & \quad+\,SmoothLLM &   2 && 0.018 & 0.004 & 0.358 && --  & --   & --   \\
      & \quad+\,AdaSteer  &   3 && 0.022 & \best{0.002} & 0.359 && --  & --   & --   \\
      & \quad+\,RESTA     &   2 && 0.013 & 0.004 & 0.351 && --  & --   & --   \\
      & \quad+\,TSA     &   \best{0} && \best{0.009} & \best{0.002} & \best{0.098} && 1.0 & 12.6 & 10.9 \\
      \addlinespace[5pt]
      & Phi-3.5-4B      &   \best{0} && 0.015 & \best{0.001} & 0.205 && --  & --   & --   \\
      & \quad+\,TSA     &   \best{0} && \best{0.009} & \best{0.001} & \best{0.098} && 1.3 & 10.2 &  8.2 \\
      \addlinespace[5pt]
      & Gemma 3-1B      &   3 && 0.023 & 0.006 & 0.354 && --  & --   & --   \\
      & \quad+\,TSA     &   \best{0} && \best{0.019} & \best{0.004} & \best{0.199} && 3.1 & 12.4 &  8.4 \\
    
    \midrule
    
    \multirow{13}{*}[-10pt]{\rotatebox[origin=c]{90}{%
      \shortstack[c]{HarmBench\\[-0.25pt]$n = 400$}}}
      & GPT-OSS-20B     &  16 && 0.046 & 0.004 & 0.947 && --  & --   & --   \\
      & \quad+\,TSA     &   \best{0} && \best{0.015} & \best{0.003} & \best{0.098} && 1.3 &  9.5 &  7.5 \\
      \addlinespace[5pt]
      & Qwen3-14B       &   8 && 0.053 & 0.020 & 0.448 && --  & --   & --   \\
      & \quad+\,TSA     &   \best{0} && \best{0.025} & \best{0.010} & \best{0.096} && 1.9 & 20.5 & 17.9 \\
      \addlinespace[5pt]
      & Llama 3.1-8B    &  17 && 0.055 & 0.004 & 0.706 && --  & --   & --   \\
      & \quad+\,SmoothLLM &  13 && 0.035 & \best{0.001} & 0.811 && --  & --   & --   \\
      & \quad+\,AdaSteer  &  22 && 0.065 & 0.017 & 0.943 && --  & --   & --   \\
      & \quad+\,RESTA     &  25 && 0.072 & 0.006 & 0.852 && --  & --   & --   \\
      & \quad+\,TSA     &   \best{0} && \best{0.009} & \best{0.001} & \best{0.100} && 1.1 & 10.7 &  9.0 \\
      \addlinespace[5pt]
      & Phi-3.5-4B      &   8 && 0.050 & 0.017 & 0.523 && --  & --   & --   \\
      & \quad+\,TSA     &   \best{0} && \best{0.020} & \best{0.006} & \best{0.097} && 1.8 & 12.0 &  9.4 \\
      \addlinespace[5pt]
      & Gemma 3-1B      &  42 && 0.102 & 0.031 & 0.846 && --  & --   & --   \\
      & \quad+\,TSA     &  \best{14} && \best{0.062} & \best{0.029} & \best{0.623} && 6.3 & 19.5 & 11.3 \\
    
    \bottomrule
    \end{tabular}
    
    \end{center}
\end{table*}

Results are reported in Table~\ref{tab:main-results}.
%
%
A useful starting point for interpretation is what the oracle actually captures. The moderation score quantifies the harmfulness of the generated text, not whether the model complied with or refused the request. A refusal that reproduces the harmful query scores high because the harmful content is present regardless of intent; a substantive but harmless answer scores low. Post-TSA responses are therefore not forced into blanket refusals but remain free to engage substantively with the prompt, with the guarantee that the content they produce will be harmless.

The method's effectiveness tracks the quality of the underlying model's safety training, not merely its parameter count. Llama 3.1 (8B) converges in a single descent step while the larger Qwen3 (14B) requires roughly two. These iteration averages are conditioned on \emph{entering the optimization loop}—on Llama 3.1 and the adversarial harmful split, only 218 of 2{,}000 prompts trigger optimization at all—so the amortized per-prompt cost on a mixed, realistic workload is substantially lower than the timing figures suggest. The sole exception is Gemma 3, for which the optimization fails to recover a fraction of harmful prompts—plausibly because either its smaller embedding dimension leaves insufficient room to locate a low-harm direction inside the cosine ball, or its safety training is too thin for any nearby embedding to decode into harmless content.

The three baselines span text-level (SmoothLLM), activation-level (AdaSteer), and embedding-level (RESTA) interventions, which we compare against the dynamic embedding optimization of TSA. The decisive difference is that \emph{zeroth-order optimization ties the harmfulness objective directly to the model's inputs}, rendering the intervention agnostic to the attack mechanism. The baselines instead rely on the brittleness of adversarial suffixes (SmoothLLM, RESTA) or on steering directions derived from a fixed calibration distribution (AdaSteer). When these assumptions are violated—e.g., when adversarial intent is embedded semantically, or when harmful queries carry no adversarial suffix at all—the baselines falter; on HarmBench, AdaSteer and RESTA produce more flagged completions than the undefended model.

The do-no-harm check on the adversarial benign split passes cleanly. TSA is applied unconditionally—it is not concerned with the prompt's intent, so no refusal bias is imposed. Early stopping further prevents any detrimental changes from being applied to the embeddings. Hence, these two properties together ensure that the model's behavior on benign prompts remains intact. On the runtime side, oracle latency rather than inference is the binding constraint, which can account for roughly 20\% of the total per-prompt runtime. This suggests that a local moderation classifier or batched API requests would be an impactful deployment-time optimization.

\subsection{Sensitivity Analysis}
\begin{figure*}[ht]
    \begin{center}
        \subfigure[Perturbation scale ($\mu$)]{
                \includegraphics[height=2.6cm]{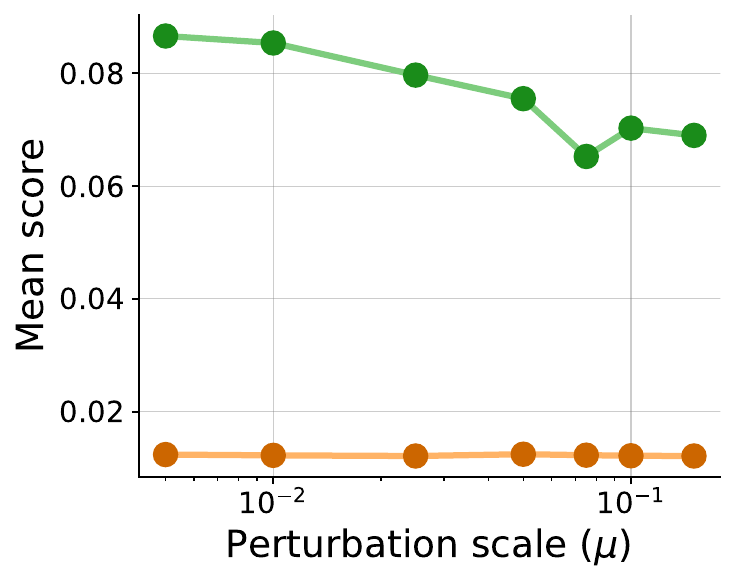}%
        }\hfill
        \subfigure[Number of samples ($N$)]{
                \includegraphics[height=2.6cm]{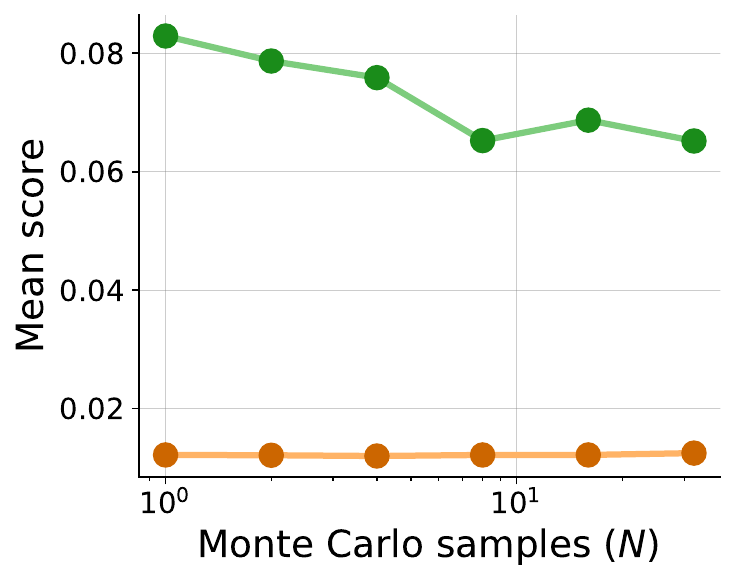}%
        }\hfill
        \subfigure[\topcap{Early stopping threshold\\(mean score)}]{
            \includegraphics[height=2.6cm]{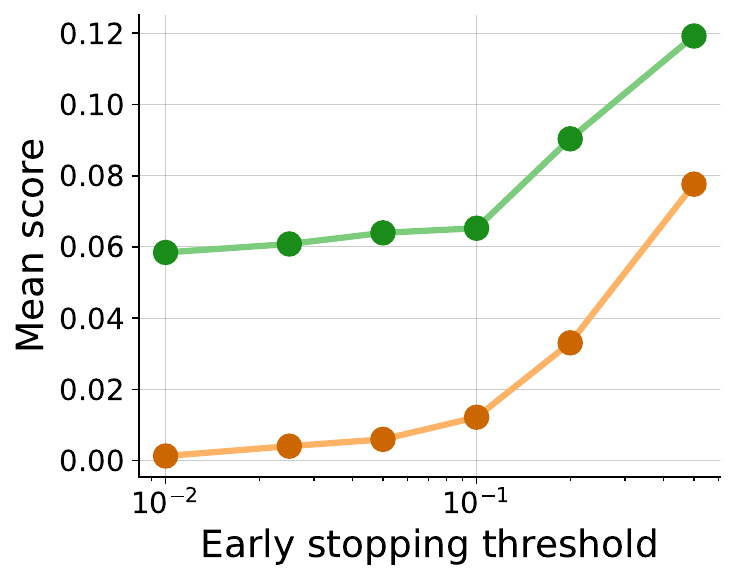}%
        }\hfill
        \subfigure[\topcap{Early stopping threshold\\(number of iterations)}]{
                \includegraphics[height=2.6cm]{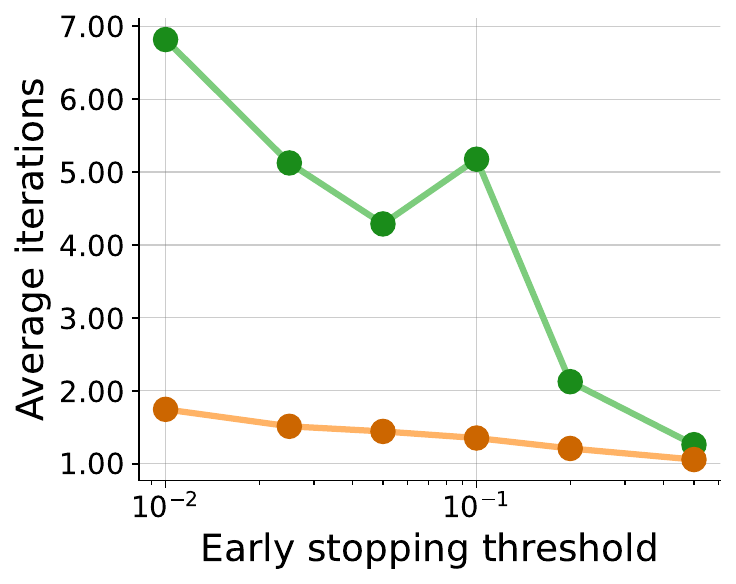}%
        }
    \caption{Sensitivity analysis on \textcolor{gemmacolor}{Gemma~3-1B} and \textcolor{llamacolor}{Llama~3.1-8B} in terms of the average mean moderation score on the adversarial harmful split ($n=2{,}000$) of the WildJailbreak benchmark. The average number of iterations to convergence is also shown for the early stopping threshold. The $x$-axis is on a logarithmic scale. \textbf{Tested values:} $\mu \in \{0.005, 0.01, 0.025, 0.05, 0.075, 0.1, 0.15\}$, $N \in \{1, 2, 4, 8, 16, 32\}$, and $\text{threshold} \in \{0.01, 0.025, 0.05, 0.1, 0.2, 0.5\}$.}
    \label{fig:sensitivity}
    \end{center}
\end{figure*}

The end-to-end calibration depends on three key variables: the Monte Carlo sample count $N$, the perturbation strength $\mu$, and the moderation threshold. The 0.1 moderation threshold is borrowed from common practice among the API developers, so we also examine how the calibration behaves as the threshold varies. Results are visualized in Figure~\ref{fig:sensitivity}.

Llama~3.1 is largely invariant to both $\mu$ and $N$ because the prompts that enter the optimization loop typically sit near a broad safe basin where even a crude descent direction suffices. This is consistent with a recent observation~\citep{subspaces} that the representations of benign and harmful prompts—once adversarially framed—lie very close to each other in the hidden space, allowing a single gradient direction to readily steer the input representation back onto the malicious manifold and elicit a more cautious response. Gemma~3's thinner alignment and lower embedding dimensionality, on the other hand, present a narrower target, explaining its greater sensitivity. Larger $N$ helps slightly, though gradient normalization ensures that only the direction of the estimate governs the update, so even a single perturbation sample can provide sufficient directional signal. Notably, $N = 8$ suffices across embedding dimensions from 1152 to 5120, despite estimator variance scaling with the ambient dimension $Td$, suggesting the safety-relevant optimization direction need not be especially precise.

Neither model displays a monotonic trend in $\mu$, confirming a mild bias--variance tradeoff across the tested range. The early stopping threshold is the only parameter with a monotonic effect (by construction), but the more informative observation is that Llama reaches a mean score of 0.001 at the 0.01 threshold in fewer than two iterations—a single gradient step typically overshoots the default threshold by a wide margin, indicating a steep moderation-score landscape near the safety boundary.

\section{Discussion}
\label{sec:discussion}

\paragraph{How can a sub-lexical intervention be so effective?}
This is a joint consequence of the regularization and the geometry of the embedding space. Gradient normalization keeps the cumulative update small in magnitude, while the vocabulary—on the order of $10^5$ points in $\mathbb{R}^d$ for $d \ge 1152$—occupies a vanishingly sparse subset of the embedding space, so each token's Voronoi cell is large relative to the step sizes involved. The perturbation therefore stays within each cell, yet the model receives genuinely different continuous inputs—inputs that share a discrete token identity but differ in the sub-lexical representation the transformer actually conditions on. These differences propagate through the network layers, producing materially different hidden states and output distributions—confirming that autoregressive generation is sensitive to continuous variation within a single token's embedding subspace.

\paragraph{Can the optimization be reversed to produce jailbreaks?}
Reversing the descent direction to maximize harmfulness failed to produce meaningfully harmful outputs in our earlier experiments. We attribute the asymmetry to the effects of safety training. The moderation scores of aligned models' default responses are typically already low, so the finite-difference numerator collapses into estimator noise and the ascent direction carries no usable signal. Safety training further shapes a broad refusal basin in which most local perturbations continue to refuse, while the narrow compliance regions exploited by known jailbreaks lie beyond the reach of small isotropic Gaussian probes. The failure of local ascent is therefore less a limitation of the optimizer than indirect evidence that safety training has rendered the neighborhood of aligned behavior geometrically inhospitable to low-budget attacks.

\paragraph{Practical profile.}
Across four of the five models evaluated, every safety-flagged response is eliminated on all benchmarks. Achieving this requires only the input word embeddings and a black-box moderation oracle—once the embeddings are in hand, the method operates fully \emph{black-box} with no access to model internals.
%
%
%
Inference engines such as vLLM~\citep{vllm} can batch the $N+1$ perturbations into a single call, reducing the overhead to a small constant factor over standard single-pass generation. The moderation oracle—the OpenAI Moderation API—is freely available, and oracle latency rather than model inference is the binding runtime constraint (Table~\ref{tab:main-results}). In effect, the method serves as a drop-in safety layer that asks nothing of the practitioner beyond the ability to run inference and call a public API, making it straightforwardly deployable in real-world scenarios.

\paragraph{Limitations.}
The optimization loop introduces per-prompt latency (typically on the order of seconds, given convergence in one or two gradient steps) which is the most visible deployment consideration. The oracle's network round-trip is the only irreducible component of this cost.
%
%
The method also requires access to the input word embeddings, a condition satisfied by any open-weight model but not by purely API-gated services. Finally, the Gemma~3 results indicate that effectiveness scales with the quality of the model's existing safety training: when alignment is thin or the embedding dimension is small, the optimizer may not easily find a nearby direction that decodes into harmless content. The method amplifies existing safety alignment; it does not fully substitute for it.
\section{Conclusion}
\label{sec:conclusion}


Embedding-level behavioral control, previously demonstrated for reducing surface-level profanity in open-ended completion models, extends to the fundamentally harder regime of safeguarding aligned models against prompts that elicit harmful content. By estimating the gradient of a black-box content-moderation oracle with respect to the prompt embeddings and descending along it, the proposed method minimizes the semantic harmfulness of generated text—without retraining, auxiliary modules, or any knowledge of prompt intent—and adapts its intervention to each input through a handful of gradient steps. Applicable to any off-the-shelf instruction-tuned model as a drop-in safety layer, the method eliminates nearly all safety-flagged completions across five models and standard red-teaming benchmarks while leaving already-safe responses intact.

Beyond the immediate safety application, our results reveal the embedding layer as a richer control surface for aligned models than its standard role as a discrete token lookup would suggest. In every case, the optimized embeddings decode back to the original prompt tokens—the perturbation is entirely sub-lexical—yet the model's output distribution shifts reliably toward safe behavior. This sensitivity to continuous variation below the granularity of discrete tokens implies that the embedding space encodes behavioral degrees of freedom invisible to any method operating at the token level. Whether sub-lexical embedding optimization generalizes as a mechanism for test-time behavioral control beyond safety—to objectives such as instruction-following or policy compliance, to other model families, or to other modalities—is a natural direction for future work.


\bibliographystyle{plainnat}
\bibliography{references}

\clearpage
\appendix
\section{Algorithm Pseudocode}
\label{app:pseudocode}

Algorithm~\ref{alg:tsa} provides the complete pseudocode for the test-time safety alignment procedure described in Section~\ref{sec:methodology}. When early stopping does not trigger, each iteration requires $N + 1$ forward passes through the language model and $N + 1$ moderation oracle queries---one base-point evaluation and $N$ perturbed evaluations---all independent and batchable into a single inference call. When the base-point score already falls below the early stopping threshold, the gradient computation is skipped entirely, reducing that iteration to a single forward pass. $\Pi_\kappa$ denotes projection onto $\{Z : \cos(Z, X_0) \ge \kappa\}$.

\begin{algorithm}[ht]
\caption{Test-Time Safety Alignment (TSA)}
\label{alg:tsa}
\begin{algorithmic}[1]

\Require Prompt $\mathcal{P}$;\; language model $f$;\; moderation oracle $h$

\Statex

\State $\mathcal{P}_{\mathrm{chat}} \gets \Call{ChatTemplate}{\mathcal{P}}$ \Comment{Format with role markers}
\State $t_{1:T} \gets \Call{Tokenize}{\mathcal{P}_{\mathrm{chat}}}$ \Comment{Token sequence $(t_1, \ldots, t_T)$}
\State $X_0 \gets \Call{Embed}{t_{1:T}}$ \Comment{Word embeddings $X_0 \in \mathbb{R}^{T \times d}$}

\Statex

\State $\Phi^* \gets \infty$;\; $X^* \gets X_0$ \Comment{Track best embedding}

\Statex

\For{$k = 0, \ldots, K - 1$}
    \State $y_k \gets f(X_k)$ \Comment{Generate response}
    \State $\Phi_k \gets h(y_k)$ \Comment{Moderation score $\Phi_k = h(f(X_k))$}
    \Statex
    \If{$\Phi_k < \Phi^*$} \Comment{Update best}
        \State $\Phi^* \gets \Phi_k$;\; $X^* \gets X_k$
    \EndIf
    \Statex
    \If{$\Phi_k < 0.1$} \Comment{Early stopping}
        \State \textbf{break}
    \EndIf
    \Statex
    \For{$i = 1, \ldots, N$} \Comment{$N$ perturbations}
        \State $U_i \sim \mathcal{N}(\mathbf{0},\, I) \in \mathbb{R}^{T \times d}$
        \State $y_i \gets f(X_k + \mu\, U_i)$
        \State $\Phi_i \gets h(y_i)$
    \EndFor
    \Statex
    \State $\displaystyle g_{k} \gets \frac{1}{N} \sum_{i=1}^{N} \frac{\Phi_i - \Phi_k}{\mu}\, U_i$ \Comment{Zeroth-order gradient estimate}
    \State $g_{k} \gets g_{k} \,/\, \lVert g_{k} \rVert_2$ \Comment{Gradient normalization}
    \State $X_{k+1} \gets X_k - \eta\, g_{k}$ \Comment{Descent step}
    \Statex
    \If{$\cos(X_{k+1},\, X_0) < \kappa$} \Comment{Cosine similarity constraint}
        \State $X_{k+1} \gets \Pi_\kappa(X_{k+1},\, X_0)$
    \EndIf
\EndFor

\Statex

\State \Return $X^*$

\end{algorithmic}
\end{algorithm}
\clearpage

\section{Experimental Details}
\label{app:experimental-details}

\begin{table*}[!htb]
    \caption{Sample counts and token-level statistics for each safety-alignment dataset. No preprocessing is applied.}   \label{tab:dataset_stats}
    \begin{center}
    \begin{tabular}{lccccc}
        \toprule
        &  & \multicolumn{4}{c}{\# Tokens} \\
        \cmidrule(lr){3-6}
        \textbf{Dataset} & \# Samples & Max & Min & Mean & Median \\
        \midrule
        WildJailbreak (adversarial benign) & 210 & 601 & 14 & 191.15 & 157 \\
        WildJailbreak (adversarial harmful) & 2,000 & 614 & 18 & 141.97 & 126 \\
        \midrule
        HarmBench (direct harmful) & 400 & 39 & 6 & 17.86 & 17 \\
        \bottomrule
    \end{tabular}
    \end{center}
\end{table*}

\subsection{Benchmarks}
\label{app:datasets}

Detailed statistics covering the number of samples and token-level properties (minimum, maximum, mean, and median) are provided in Table~\ref{tab:dataset_stats}.

The selected benchmarks were originally designed for red-teaming evaluation; we repurpose them as sources of prompts that span a range of conditions under which aligned models may produce harmful content. WildJailbreak was additionally proposed for safety training and validation tasks. We also considered its sister dataset, WildGuardMix~\citep{wildgm}, from the same developer, which is designed primarily for \emph{moderation}—that is, teaching models to refuse harmful queries appropriately. Since WildGuardMix is derived from WildJailbreak, we proceeded with the latter.

\paragraph{Direct Harmful (HarmBench).}
The authors used GPT-4~\citep{gpt4} to distill the acceptable use policies of major AI companies into a set of guidelines, and then manually designed behaviors that violate laws or norms while filtering out ``dual-intent'' behaviors with plausible benign uses. They prioritized \emph{differential harm}—behaviors that LLMs can enable beyond what a person could readily obtain through a search engine—particularly for contextual and multimodal cases.

\paragraph{Harmful Injections (WildJailbreak).}
Harmful requests framed adversarially as prompt injections in more convoluted and stealthy forms. The authors' WildTeaming framework was applied to rewrite direct harmful queries using 2--7 randomly sampled in-the-wild jailbreak tactics, with Mixtral-8$\times$7B~\citep{mixtral} and GPT-4 serving as the rewriters. After filtering out low-risk and off-topic prompts, each adversarial prompt was paired with the refusal response of its direct counterpart, yielding 82{,}728 items.

\paragraph{Benign Injections (WildJailbreak).}
Prompt injections that resemble jailbreaks but carry no harmful intent. \citet{wildjb} generated 78{,}710 such prompts with WildTeaming from direct benign queries, using GPT-3.5~\citep{gpt3} to produce the direct prompts.

\subsection{Hyperparameters}
\label{app:hyperparams}

We swept $\mu$ and the learning rate $\eta$ on a small representative subset of 75 samples consisting mostly of adversarial harmful prompts, together with a small number of adversarial benign and direct harmful prompts as a sanity check. For each model, we typically grid-searched four to five values of $\mu$ and six to seven values of $\eta$, yielding 24 to 35 configurations in total. Most configurations with aggressive learning rates performed similarly, differing only modestly in the average number of iterations. The final tuned values are listed in Table~\ref{tab:tuned_params}. 

\begin{table}[!htb]
    \centering
    \caption{Model configurations and tuned TSA hyperparameters, along with the embedding (i.e., hidden) dimensionality and the number of hidden layers. For each model, the same hyperparameters are used across all benchmarks. A moderation threshold of 0.1 is used throughout.}
    \label{tab:tuned_params}
        \begin{tabular}{@{}lccccc@{}}
        \toprule
        \textbf{Parameter}           & GPT-OSS & Qwen3 & Llama 3.1 & Phi-3.5 & Gemma 3 \\
        \midrule
        Size                         & 20B     & 14B   & 8B        & 4B      & 1B      \\
        Dimensionality $d$           & 2880    & 5120  & 4096      & 3072    & 1152    \\
        \# Layers                    & 24      & 40    & 32        & 32      & 26      \\
        \midrule
        \# Monte Carlo samples       & 8       & 8     & 8         & 8       & 8       \\
        Perturbation scale $\mu$     & 0.01    & 0.05  & 0.1       & 0.05    & 0.075   \\
        Step size $\eta$              & 1.5     & 2.5   & 1.3       & 2.5     & 1.75    \\
        Cosine similarity threshold  & 0.2     & 0.2   & 0.2       & 0.2     & 0.2     \\
        \bottomrule
    \end{tabular}
\end{table}

\subsection{Implementation}
\label{app:implementation}

Since our method operates purely through forward passes, we use vLLM~\citep{vllm} for both the test-time optimization and the base-model generations. Through continuous batching across prompts and decoding steps, vLLM enables concurrent processing while sharing model execution and KV-cache management, making the overhead of batching negligible relative to single-input inference. Our test-time procedure therefore reduces to a small number of standard forward passes, effectively proportional to the number of optimization iterations in practice.

For the baselines, SmoothLLM is implemented on top of vLLM (the same inference backend used by our method) and applies random character swaps at a 10\% perturbation rate across $k{=}10$ copies per prompt, with the returned completion selected by majority vote over a substring-based jailbreak judge. For AdaSteer, per-layer Rejection Direction (RD) and Harmfulness Direction (HD) steering vectors are extracted via difference-in-means over 256 harmful--benign contrast pairs drawn from the WildJailbreak training split. At inference time, adaptive per-input steering coefficients are computed from piecewise-linear maps evaluated at layer~8 (RD) and layer~13 (HD) for Llama 3.1-8B-Instruct, following the original hyperparameters reported for this architecture. RESTA adds isotropic Gaussian noise to the user-content token embeddings; the original paper uses $\sigma = 0.05$ for Llama 2-7B, but we found $\sigma = 0.08$ to work better with Llama 3.1-8B. The perturbed embeddings yield $k{=}10$ noisy copies, which are aggregated via majority vote over a smoothed prefix of length $l{=}10$, after which generation proceeds greedily from the agreed-upon prefix. AdaSteer and RESTA require embedding- or activation-level model access and therefore use HuggingFace \texttt{transformers}~\citep{transformers}, whereas SmoothLLM operates at the character level.

\subsection{Compute}
\label{app:compute}

All experiments were conducted on a single local machine equipped with two NVIDIA RTX A6000 GPUs (49\,GiB each). Models were served via vLLM with tensor parallel size~2 and GPU memory utilization set to 95\%. Per-prompt wall-clock timing under this configuration, including and excluding moderation API latency, is reported in Table~\ref{tab:main-results}.

\end{document}